\documentclass[10pt, a4paper]{article}

\usepackage[]{lrec-coling2024} 

\usepackage{times}
\usepackage{latexsym}
\usepackage{graphicx}
\usepackage{multirow}

\usepackage[T1]{fontenc}

\usepackage[utf8]{inputenc}

\usepackage{microtype}

\usepackage{acronym}
\acrodef{LLM}[LLM]{Large Language Model}
\acrodef{NLP}[NLP]{Natural Language Processing}
\acrodef{GPT}[GPT]{Generative Pre-trained Transformer}
\acrodef{BERT}[BERT]{Bidirectional Encoder Representations from Transformers}
\acrodef{T5}[T5]{Text-To-Text Transfer Transformer}
\acrodef{CLS}[CLS]{Classification}
\acrodef{RE}[RE]{Relation Extraction}
\acrodef{QA}[QA]{Question Answering}
\acrodef{NLI}[NLI]{Natural Language Inference}
\acrodef{STS}[STS]{Semantic Textual Similarity}
\acrodef{NER}[NER]{Named Entity Recognition}
\acrodef{FSL}[FSL]{Few-Shot Learning}

\usepackage{numprint}
\npdecimalsign{.}
\nprounddigits{2}

\title{A Zero-shot and Few-shot Study of Instruction-Finetuned Large Language Models Applied to Clinical and Biomedical Tasks}

\name{Yanis Labrak{\normalfont $^*$\textsuperscript{1,2}}, Mickael Rouvier{\normalfont $^*$\textsuperscript{1}} and Richard Dufour{\normalfont $^*$\textsuperscript{1,3}}} 

\address{\textsuperscript{1}LIA, Avignon Université \hspace{4mm} \textsuperscript{2}Zenidoc \\ \textsuperscript{3}Nantes Université, École Centrale Nantes, CNRS, LS2N, UMR 6004, F-44000 Nantes, France  \\
         \{first.last\}@\{univ-avignon.fr, univ-nantes.fr\} \\}

\abstract{
The recent emergence of Large Language Models (LLMs) has enabled significant advances in the field of Natural Language Processing (NLP). While these new models have demonstrated superior performance on various tasks, their application and potential are still underexplored, both in terms of the diversity of tasks they can handle and their domain of application. In this context, we evaluate four state-of-the-art instruction-tuned LLMs (ChatGPT, Flan-T5 UL2, Tk-Instruct, and Alpaca) on a set of 13 real-world clinical and biomedical NLP tasks in English, including named-entity recognition (NER), question-answering (QA), relation extraction (RE), and more. Our overall results show that these evaluated LLMs approach the performance of state-of-the-art models in zero- and few-shot scenarios for most tasks, particularly excelling in the QA task, even though they have never encountered examples from these tasks before. However, we also observe that the classification and RE tasks fall short of the performance achievable with specifically trained models designed for the medical field, such as PubMedBERT. Finally, we note that no single LLM outperforms all others across all studied tasks, with some models proving more suitable for certain tasks than others.
 \\ \newline  }

\begin{document}

\maketitleabstract

\section{Introduction}

Medical domain is currently benefiting greatly from significant progress in Natural Language Processing (NLP), thanks to the availability of massive textual databases and the use of deep learning techniques that allow for more efficient exploitation of this data. Traditionally, the approach involved training a generic masked language model (MLM) and then adapting it to a specific domain or task, such as BERT models~\cite{devlin-etal-2019-bert}. However, the latest approaches aim to develop Large Language Models (LLMs) that can directly process a wide range of NLP tasks and domains. They can then handle tasks such as classification or entity extraction, as well as more complex generative tasks like machine translation or question-answering.

While there is clear enthusiasm for LLMs among both scientists and the general public, the evaluation of these models, also known as foundation models, is still in its infancy. The initial evaluations demonstrate the usefulness of these models in performing various NLP tasks, including classification and generation tasks on general domains~\cite{liu2023gpteval,bang2023multitask}. However, in the medical field, these models have been evaluated to a lesser extent, often on a limited number of tasks~\cite{rehana2023evaluation, chen2023evaluation,lamichhane2023evaluation, singhal2022large, chowdhery2022palm}. This is mainly due to the scarcity of tasks and data, particularly sensitive data that is difficult to obtain, compared to other fields.


To evaluate how well LLMs encode medical knowledge and to demonstrate their capabilities in specific domains, a wide range of tasks that closely resemble real-world applications and require appropriate medical knowledge and expert reasoning were considered. Unlike other studies~\cite{fries2022bigbio, Medical2021} that have compared performances of these models using automatic metrics (BLUE~\cite{10.3115/1073083.1073135}, ROUGE~\cite{lin-2004-rouge} or BertScore~\cite{Zhang2020BERTScore}) or only accuracy on a small set of tasks, we decide to showcase their relevance in various evaluation contexts by using more commonly used metrics (Accuracy and F1) which are allowing a fair direct comparison with BERT-based models. In overall, we curate a collection comprising 13 real-world medical tasks, including classification (CLS), question-answering (QA), relation extraction (RE), natural language inference (NLI) and named-entity recognition (NER). 

The main contributions of the paper are:

\begin{itemize}
\item Evaluation of four state-of-the-art instruction-tuned models (ChatGPT, Flan-T5 UL2, Tk-Instruct, and Alpaca) on a broad range of medical tasks in English language beyond those typically addressed by generative models.

\item Assessment of the ability of the studied language models to perform zero- and few-shot inference and comparison of their performance on the tasks with that of a fine-tuned PubMedBERT model.

\item Introduction of a novel method called Recursive Chain-of-Thought (RCoT) that enables performing the NER task on all types of LLMs thanks to the use of a prompt sequentially enriched to mimic human reasoning.






\end{itemize}




\section{Related work}
\label{sec:related-work}

We first introduce the concept of Large Language Models (LLMs) and their limitations (Section~\ref{sec:language_modeline}). Next, we present the concept of instruction-tuning (Section~\ref{s:intruc-tuning}). Finally, we describe our few-shot learning strategy with prompts (Section~\ref{sec:instruction_based_strategies}).


\subsection{Large Language Models (LLMs)}
\label{sec:language_modeline}

While classical language models like BERT are efficient across various NLP tasks and trained on substantial amounts of unannotated textual data, they still necessitate a significant quantity of annotated data to excel in specific tasks, such as NER, NLI, and RE. Moreover, these models encounter challenges when attempting to generalize their knowledge to other languages or domains after being adapted to a particular task and context~\cite{peng2021domain}. Collecting such data for any scenario can be costly, demanding highly skilled annotators and giving rise to privacy concerns.

Recently, LLMs have brought additional performance improvements, especially in generative tasks. These models are composed of billion of parameters and trained on gigantic amounts of data, from various natures, domains and languages~\cite{gao2020pile, JMLR:v21:20-074, OrtizSuarezSagotRomary2019}. Previous studies have demonstrated in particular that this gigantic number of parameters associated with this massive data allowed the fine modeling of the language, making it possible to achieve this level of performance~\cite{zhang2022opt, black-etal-2022-gpt, hoffmann2022training, smith2022using}.


New approaches leveraging the generative capabilities of LLMs have aimed to align them with instructions~\cite{NEURIPS2022_b1efde53} (see Section~\ref{s:intruc-tuning}), thereby enhancing their capacity to handle a multitude of NLP tasks in multiple languages using zero-shot or few-shot learning~\cite{bang2023multitask}.



\subsection{Instruction Tuning} 
\label{s:intruc-tuning}

\citet{efrat2020turking} and~\citet{mishra-etal-2022-cross} propose the instruction paradigm, in which models can learn new tasks based on natural language instructions only. These instructions are given as inputs to the models, describing how they should behave, what we expect from them, and on which information they can base their thinking on. \citet{wang-etal-2022-super} introduced the first large-scale instruction benchmark called \textsc{Super-NaturalInstructions}, by collecting crowdsourced instructions based on an existing set of 1600+ NLP datasets and converting them into a uniform format. \citet{sanh2022multitask} and~\citet{wei2022finetuned} further extend the adoption of instructions by suggesting instruction tuning, in which a LLM is trained on many natural language instructions with the aspiration that it will generalize to new, unseen instruction tasks. \citet{chung2022scaling} advance instruction tuning by scaling the number of tasks, scaling the model size, and introducing the concept of chain-of-thought~\cite{wei2022chain}, while~\citet{NEURIPS2022_b1efde53} propose a reinforcement learning approach for instruction tuning and human feedback.



\subsection{Few-shot Learning with prompts}
\label{sec:instruction_based_strategies}

During inference, a few examples of the task are given to the model as conditioning, without updating its weights. These examples usually comprise an instruction, context, and desired completion (e.g., a premise, hypothesis, and corresponding label for the NLI task). The few-shot technique involves presenting the model with $k$ examples of context and completion, followed by a final example of context, for which the model should provide the completion. The value of $k$ typically ranges from 3 to 100, which depends on the number of examples that can fit within the model's context window (for instance, Flan-UL2 has a context window of 2,048 tokens).
\section{Experimental Protocol}
\label{sec:experimental-protocol}

In this section, we describe the models utilized and the datasets used to benchmark the various models.

\subsection{Studied Models}

Our evaluation involves four distinct generic LLMs (ChatGPT, Flan-UL2, Tk-Instruct and Alpaca) and a specific biomedical masked language model (PubMedBERT) for comparison purposes.


\paragraph{Flan-T5 UL2} abbreviated to Flan-UL2, is an encoder-decoder model based on UL2 20B parameters model~\cite{tay2023ul2} and was fine-tuned using the Flan instruction tuning tasks collection~\cite{chung2022scaling}.




\paragraph{Tk-Instruct} is based on the T5 encoder-decoder model~\cite{JMLR:v21:20-074} and has been fine-tuned on the 1,600+ NLP tasks from the \textsc{Super-NaturalInstructions} dataset~\cite{wang-etal-2022-super}. In our study, we chose the 3B parameter setting, since our preliminary comparison with Flan-T5-XL~\cite{https://doi.org/10.48550/arxiv.2210.11416} using the 3B parameter setting showed that Tk-Instruct performed better on QA tasks, which is considered to be one of the most suited tasks for LLMs.


\paragraph{ChatGPT} is built upon GPT-3.5 Turbo, fine-tuned with a set of proprietary instructions, and continuously refined through reinforcement learning from human feedback (RLHF) techniques. Access to its weights is restricted, and the model can only be accessed via a paid API. These restrictions raise privacy concerns regarding its application in medical contexts, and it cannot ensure that the evaluated data has not been previously encountered.


\paragraph{Stanford Alpaca} is built upon LLaMA with 7B parameters~\cite{touvron2023llama} and utilizes a dataset of 52K instructions, which were automatically generated in the style of self-instruct using OpenAI's text-davinci-003 model~\cite{wang2022selfinstruct}. Due to its base model and data sources, it is exclusively intended for academic research purposes and non-commercial use.


\paragraph{PubMedBERT} is a biomedical-specific BERT-based model with 110M parameters~\cite{10.1145/3458754}. It was trained entirely from scratch on the 3.1 billion words of the PubMed corpus. We chose it as our baseline for comparison with the zero-shot and few-shot performance of generative models.

\subsection{Downstream evaluation tasks}

We conducted an evaluation of the models' capabilities by encompassing the test set of the 13 diverse tasks listed in Table~\ref{table:tasks}. These tasks were chosen to facilitate a comprehensive assessment spanning both clinical and biomedical domains, including tasks suitable for both generative and classical model evaluations.

\begin{table}[!htb]
\tiny
\centering
\setlength\tabcolsep{6.5pt}
\setlength\extrarowheight{1.3pt}
\begin{tabular}{|llccc|}
\hline
\textbf{Task} & \textbf{Dataset} & \textbf{Eval} & \textbf{Metric}  & \textbf{Reference} \\
 \hline
 
\multirow{4}{*}{CLS} & HoC & Test & F1-measure & \citetlanguageresource{DBLP:journals/bioinformatics/BakerSGAHSK16}  \\
& LitCovid & Test & F1-measure & \citetlanguageresource{chen2021overview}    \\
& PubHealth & Test & Accuracy  & \citetlanguageresource{kotonya2020explainable}  \\
& N2C2 2006 Smokers  & Test & Accuracy & \citetlanguageresource{uzuner2008identifying}  \\
\hline

\multirow{4}{*}{QA} & BioASQ 7b & Test & Accuracy  & \citetlanguageresource{tsatsaronis2015overview}  \\
& MedMCQA & Dev & Accuracy  & \citetlanguageresource{pmlr-v174-pal22a}   \\
& SciQ & Test & Accuracy  & \citetlanguageresource{welbl-etal-2017-crowdsourcing}  \\
& Evidence Inference 2.0 & Test & Accuracy  & \citetlanguageresource{deyoung-etal-2020-evidence}  \\
\hline

\multirow{1}{*}{RE} & GAD & Test & Accuracy  & \citetlanguageresource{Bravo2015}  \\
\hline

\multirow{2}{*}{NLI} & SciTail  & Test & Accuracy  & \citetlanguageresource{scitail}  \\  & MedNLI & Test & Accuracy  & \citetlanguageresource{https://doi.org/10.13026/c2rs98}  \\  
\hline

\multirow{2}{*}{NER} & BC5CDR & Test & F1-measure & \citetlanguageresource{DBLP:journals/biodb/LiSJSWLDMWL16}  \\
& NCBI-disease & Test & F1-measure & \citetlanguageresource{Dogan2014NCBIDC}  \\




\hline
\end{tabular}
\caption{List of evaluation tasks and their metrics. CLS: Classification, QA: Question Answering, RE: Relation Extraction, NLI: Natural Language Inference, NER, Named-Entity Recognition. }
\label{table:tasks}
\end{table}

\subsection{Evaluation of generative outputs}

Evaluating the outputs of generative models presents a challenge due to their free-text nature, which may not necessarily conform to a predefined set of classes. Instead, we are confronted with noisy outputs that may contain correct answers. To address this challenge, we manually developed parsing scripts tailored to each task and model, aligning them with their respective output styles. This approach enables us to capture most of the answers and compute metrics that can be compared with our baseline model (PubMedBERT).

 









\begin{table*}[!htb]
\resizebox{\textwidth}{!}{%
\setlength\extrarowheight{1.7pt}
\centering
\begin{tabular}{|cl||cc|cc|cc|cc||c|}
\hline
\multirow{2}{*}{\textbf{Task}} & \multirow{2}{*}{\textbf{Dataset}}  & \multicolumn{2}{c|}{\textbf{ChatGPT}}  & \multicolumn{2}{c|}{\textbf{Flan-UL2}} & \multicolumn{2}{c|}{\textbf{Tk-Instruct}} & \multicolumn{2}{c||}{\textbf{Alpaca}} & \multirow{2}{*}{\textbf{PubMedBERT}} \\
& & \textbf{zero-shot} & \textbf{5-shot} & \textbf{zero-shot} & \textbf{5-shot} & \textbf{zero-shot} & \textbf{5-shot} & \textbf{zero-shot} & \textbf{5-shot} &  \\
\hline

\multirow{4}{*}{CLS} & HoC & \underline{\numprint{62.24}} & \numprint{38.34} & \numprint{56.36} & \numprint{54.86}  & \numprint{50.77} & \numprint{25.48} & \numprint{1.21} & \numprint{38.78}   &  \textbf{\numprint{82.75}} \\

& LitCovid & \numprint{67.20}  & \underline{\numprint{72.77}} & \numprint{51.48} & \numprint{46.95}  & \numprint{36.42} & \numprint{57.49}  & \numprint{1.58}  & \numprint{64.09} & \textbf{\numprint{90.60}} \\

& PubHealth  & \numprint{63.20} & \numprint{66.29} & \underline{\numprint{72.46}} & \numprint{50.53}  & \numprint{53.70} & \numprint{66.04} & \numprint{52.80} & \numprint{55.64} &  \textbf{\numprint{75.39}}   \\

& N2C2 2006 Smokers & NaN  &  NaN  & \numprint{22.12}  & \underline{\numprint{42.31}}  & \numprint{16.35} & \numprint{37.50} &  \numprint{10.57} & \numprint{31.73}     &  \textbf{\numprint{60.58}}   \\

\hline

\multirow{4}{*}{QA} & BioASQ 7b & \numprint{89.24} & \textbf{\numprint{92.03}}  & \numprint{90.97} & \underline{\numprint{91.64}}  & \numprint{88.09} & \numprint{86.36}  & \numprint{79.05} & \numprint{79.82}   & \numprint{73.39} \\

& MedMCQA & \underline{\numprint{48.91}} & \textbf{\numprint{56.37}} & \numprint{41.05} & \numprint{43.34}    & \numprint{33.85} & \numprint{33.18} & \numprint{24.91} & \numprint{29.50} &  \numprint{38.15} \\

& SciQ & \underline{\numprint{90.10}} & \textbf{\numprint{93.50}} & \numprint{87.00} & \numprint{88.40}  & \numprint{55.30} & \numprint{47.00}  & \numprint{24.90} & \numprint{36.80}    &  \numprint{74.20}  \\

& Evidence Inference 2.0 & \numprint{59.98} & \numprint{63.83}  & \underline{\numprint{66.45}}   & \numprint{65.06} 
 & \numprint{41.33} & \numprint{38.79} & \numprint{32.49}  & \textbf{\numprint{94.18}} & \numprint{65.47}  \\

 \hline

\multirow{1}{*}{RE} &  GAD & \numprint{47.75} & \numprint{52.25} & \numprint{49.81} &  \numprint{53.37}  & \numprint{48.88} &  \underline{\numprint{57.87}} & \numprint{51.12}  &   \numprint{57.68} & \textbf{\numprint{79.78}}  \\

\hline

\multirow{2}{*}{NLI} & SciTail & \numprint{73.57} & \numprint{65.62} & \textbf{\numprint{93.51}}  & \underline{\numprint{92.66}} & \numprint{57.53} & \numprint{71.31} & \numprint{39.60} & \numprint{40.26} & \textbf{\numprint{93.51}}   \\  

& MedNLI & NaN & NaN & \numprint{77.00}  & \underline{\numprint{79.18}} & \numprint{33.19} & \numprint{34.81}& \numprint{33.47} & \numprint{34.45}  & \textbf{\numprint{83.76}}    \\

\hline

\multirow{2}{*}{NER} & BC5CDR & \numprint{92.12} &  \underline{\numprint{93.12}} & \numprint{68.26}  &  \numprint{83.32} &  \numprint{84.54} 
 & \numprint{83.23}  & \numprint{82.11}& \numprint{84.07} &  \textbf{\numprint{97.65}}    \\

& NCBI-disease & \numprint{90.97} & \underline{\numprint{92.27}} & \numprint{90.75} & \numprint{87.65}  & \numprint{87.91} &  \numprint{87.50} & \numprint{11.58} & \underline{\numprint{92.27}} & \textbf{\numprint{98.72}} \\

\hline
\end{tabular}
}
\caption{0- and 5-shot versus finetuning evaluation on clinical and biomedical tasks. Bold values are the highest scores obtained for the task and in underlined the seconds ones. Not allowed experiments are replaced by \texttt{NaN}.}
\label{table:all-results}

\end{table*}


\subsection{Instruction Format} Previous studies~\cite{wei2022chain, jung-etal-2022-maieutic, mishra-etal-2022-reframing} have demonstrated the effectiveness of using task-specific prompts for each model. Consequently, we chose to construct the input instruction prompt by concatenating three elements: (1) an instruction that outlines the task, describes the nature of the data, and specifies our expectations from the model, (2) the input argument, which provides essential information for the task, and (3) the constraints on the output space, which guide the model during output generation. Lastly, the output serves as a reference point during the few-shot strategy evaluation. More information about the instruction formats in Appendix \ref{sec:instructions-examples}.


\subsection{Few-shot Examples using Semantic Retriever}\label{sec:few-shot} To enhance few-shot performance compared to randomly sampled examples, we introduced an additional retrieval module based on Sentence-Transformers~\cite{reimers-2019-sentence-bert}. The objective is to identify the $k$ most semantically similar examples from the training set. To accomplish this, we first populate a vector space with sentence representations of each individual instruction prompt from the training set, obtained using a pre-trained and fixed PubMedBERT~\cite{10.1145/3458754} model. Subsequently, we compute the cosine distance between the query of the current test instance and all the elements within the vector space to retrieve the top $k$ closest examples. In our case, we set the value of $k$ to 5.

\subsection{Recursive Chain-of-Thought} We performed NER using two inference methods. The first one is based on the method introduced by~\citet{ye2023comprehensive} and can only be applied using ChatGPT. It consists of giving the model a sequence of words separated by double vertical bars for word separation and single vertical bars for the separation between words and labels. For the second method, we introduce a method called Recursive Chain-of-Thought (RCoT). It is very close to human reasoning and works for all the generative models we have tried. It is derived from the Chain-of-Thought (CoT) concept~\cite{wei2022chain} and the work of~\citet{wang-etal-2022-super}. It involves iterating over the sequence of tokens and giving the current state of the prediction as input to the model, asking for the generation of the label of the $N^{th}$ token. This method guarantees an entity for each token of the sequence and prevents forgotten tokens during generation. However, the only drawback we have been able to identify with this method is its very high computation cost due to its $\mathcal{O}^N$ complexity, with $N$ being the number of tokens in the sequence, compared to the method used for ChatGPT, which performs at $\mathcal{O}^1$ complexity.










\section{Results and Discussions}
\label{sec:results-discussions}

Table~\ref{table:all-results} reports performance obtained on each task by the studied LLMs in zero- and few-shot scenario, as well as PubMedBERT fine-tuned. Results are reported by taking the best run out of four.





\paragraph{Zero-shot scenario}

Compared to PubMedBERT, the zero-shot scenario results show a clear deficit for the generative models on all the tasks except for QA, in which LLMs obtain better performance. ChatGPT and Flan-T5 UL2 are particularly perform better than Tk-Instruct and Alpaca on average, except for GAD dataset (RE task) for which Alpaca reaches the best performance. We can also observe extremely poor performance from Alpaca in zero-shot scenario on the two CLS tasks (HoC and LitCovid). These low scores are attributed to the model generating hallucinated responses, including the label \textit{evading growth suppressors} across the entire test set of HoC. However, this behavior does not appear to occur in the few-shot scenario, where the model appears to comprehend our expectations.

\paragraph{Few-shot capabilities}

Unlike the zero-shot scenario, the few-shot inference (5-shots in our experiments) shows impressive behavior. The biggest absolute gains are obtained using Alpaca, which seems to perform much better in few-shot scenarios on all tasks. We suspect this behavior to be correlated with Alpaca's training data, which does not contain many similar instructions for the tasks we are trying to tackle, allowing it to better understand what we are asking when confronted with dissimilar examples. ChatGPT also benefits from the additional knowledge to further improve the already good results, especially on QA tasks. Flan-T5 UL2 appears to be less affected by the additional context overall, except for the BC5CDR and N2C2 2006 Smokers tasks.

\section{Conclusion}
\label{sec:conclusion}

In this study, we have demonstrated that generic LLMs are capable of capturing medical knowledge and performing exceptionally well in zero- and few-shot scenarios, despite having no prior exposure to the tasks. Although open-source models such as Flan-T5 UL2 are gradually approaching their closed-source counterparts, such as ChatGPT, their performance still lags behind. We suggest that developing domain-specific models, fine-tuned on a diverse set of tasks and specialized instruction prompts, could help bridge the gap with more robust and performant proprietary models. We also note that domain-specific BERT models remain a viable option, but require a significant amount of data for fine-tuning on targeted languages and tasks. However, BERT-based models offer much lower computational costs compared to LLMs, which could be a significant obstacle to developing models in the healthcare domain.

\section{Acknowledgements}
\label{sec:Acknowledgements}

This work was performed using HPC resources from GENCI-IDRIS (Grant 2022-AD011013061R1). This work was financially supported by ANR MALADES (ANR-23-IAS1-0005) and Zenidoc.

\section{Limitations}

Through all the experiments we conducted, we have observed that Large Language Models (LLMs) trained based on instructions often exhibit sensitivity to the specific wording used as input, which can influence their ability to generate correct outputs. This finding may not come as a surprise, as LLMs are well-documented to be highly responsive to the prompts they receive, whether in zero-shot or few-shot settings [cite relevant sources]. However, it frequently necessitates tailoring the prompts to suit the models and tasks, or even mapping the classes to more suitable ones. This sensitivity may stem from the limited diversity in the collections of instructions used for their training. One of the primary limitations is our inability to guarantee that the ChatGPT model has not encountered the evaluation data during its training, potentially introducing bias into the results. Similarly, Flan-T5 UL2 and Tk-Instruct have been trained on a broad spectrum of tasks, which could result in the model being exposed to similar or identical data if overlaps are not identified. As a result, we cannot ensure that the training data for certain tasks has never been seen before.

\nocite{}
\section{Bibliographical References}\label{sec:reference}

\bibliographystyle{lrec-coling2024-natbib}
\bibliography{lrec-coling2024-example}

\section{Language Resource References}
\label{lr:ref}
\bibliographystylelanguageresource{lrec-coling2024-natbib}
\bibliographylanguageresource{languageresource}

\newpage
\onecolumn
\appendix

\section*{Appendices}
\label{sec:appendix}

\section{Instructions Formats}
\label{sec:instructions-examples}

The following sections are giving example of prompts used for training and inference for organized by tasks.

\subsection{Named-Entities Recognition}

\subsubsection{Method 1}

\begin{table}[h]
\small
\hrule  \vspace{3mm}
Prompts \\ \hrule  \vspace{3mm}

\textbf{Instruction:} Do named-entity recognition task for the given text using the categories in candidate list, output using the format as “Word1|Category||Word2|Category||Word3|Category”

\textbf{Candidate list:} \textit{O}, \textit{B-Disease} or \textit{I-Disease}

\textbf{Text:} Identification|Category || of|Category || APC2|Category || ,|Category || a|Category || homologue|Category  || of|Category || the|Category || adenomatous|Category || polyposis|Category || coli|Category || tumour|Category
|| suppressor|Category || .|Category 

\textbf{Output:} \\   \vspace{1mm}

\textbf{Instruction:} You are a healthcare named-entity recognition expert system and we are giving you a sequence of words that you have to labelized using the following output format 'Word1|Label||Word2|Label||Word3|Label' 

\textbf{Labels:} \textit{O}, \textit{B-Disease} or \textit{I-Disease} 

\textbf{Unfilled sequence:} Identification|Label||of|Label||APC2|Label||,|Label||a|Label||homologue|Label||of|Label
||the|Label
||adenomatous|Label||polyposis|Label||coli|Label||tumour|Label||suppressor|Label||.|Label 

\textbf{Constraints:} The answer must be one and only one of the given labels. 

\textbf{Output:} \\   \vspace{1mm}

\textbf{Instruction:} As a healthcare named-entity recognition expert, your job is to label a sequence of words provided to you using the following format: 'Word1|Label||Word2|Label||Word3|Label'. Your goal is to identify all the named entities in the given text. The available labels for this task are: \textit{O}, \textit{B-Disease} or \textit{I-Disease} 

\textbf{Input:} Identification|Label||of|Label||APC2|Label||,|Label||a|Label||homologue|Label||of|Label||the|Label
||adenomatous|Label ||polyposis|Label||coli|Label||tumour|Label||suppressor|Label||.|Label 

\textbf{Output:} \\   \vspace{1mm}

\vspace{3mm}
\caption{Sample of three instructions used for the named-entities recognition task with ChatGPT.}
\label{table:example-ner-instructions}
\end{table}

\newpage

\subsubsection{Method 2 - Recursive Chain-Of-Thought (RCoT)}

\begin{table}[h]
\small
\hrule  \vspace{3mm}
Prompt - Recursive Chain-Of-Thought (RCoT) \vspace{-2mm}
\\ \hrule  \vspace{1mm}

\textbf{Instruction:} You are a highly intelligent and accurate healthcare domain Named-entity recognition (NER) system. You are tasked to do Named-entity recognition (NER) for 'disease' and 'none' only, please generate the appropriate label.

\textbf{Constraints:} You can choose only one label from: \textit{none} or \textit{disease}.

\textbf{Examples:} \\   \vspace{-2mm}

\textbf{Example 1 : } Mutations|none|| at|none|| the|none|| ataxia|disease|| -|disease|| telangiectasia|disease|| locus|none|| and|none|| clinical|none|| phenotypes|none|| of|none|| A|disease|| -|disease|| T|disease|| patients|none|| .|none \\   \vspace{-2mm}

\textbf{Example 2 : } Splicing|none|| defects|none|| in|none|| the|none|| ataxia|disease|| -|disease|| telangiectasia|disease|| gene|none|| ,|none|| ATM|none|| :|none|| underlying|none|| mutations|none|| and|none|| consequences|none|| .|none \\   \vspace{-2mm}

\textbf{Example 3 : } Somatic|none|| mutations|none|| in|none|| the|none|| BRCA1|none|| gene|none|| in|none|| sporadic|disease|| ovarian|disease|| tumours|disease|| .|none \\   \vspace{-2mm}

\textbf{Example 4 : } Malignant|disease|| neoplasms|disease|| in|none|| the|none|| families|none|| of|none|| patients|none|| with|none|| ataxia|disease|| -|disease|| telangiectasia|disease|| .|none \\   \vspace{-2mm}

\textbf{Example 5 : } Founder|none|| mutations|none|| in|none|| the|none|| BRCA1|none|| gene|none|| in|none|| Polish|none|| families|none|| with|none|| breast|disease|| -|disease|| ovarian|disease|| cancer|disease|| .|none \\  \vspace{-2mm} 

\textbf{Considering the sentence :} Clustering of missense mutations in the ataxia - telangiectasia gene in a sporadic T - cell leukaemia . \\   \vspace{-2mm}

\textbf{And considering your precedents predictions : } Clustering|none|| of|none|| missense|none|| mutations|none|| in|none|| the|none|| ataxia|disease|| -|disease|| telangiectasia|disease|| gene|none|| in|none|| a|none|| sporadic|disease|| T|disease|| -|disease|| cell|disease|| leukaemia|Label \\   \vspace{-2mm}

\textbf{Input :} The label of « leukaemia » at the position 17 of the sentence is ?

\textbf{Output: } 

\caption{Example of a 5-shot Recursive Chain-Of-Thought (RCoT) instruction used for the named-entities recognition task of NCBI Disease dataset.}
\label{table:example-ner-instructions-SCoT}
\end{table}


\section{Multiple-choice question answering}

\subsection{Method 1 - One-shot}

\begin{table}[h]
\small
\hrule  \vspace{3mm}
Prompt \\ \hrule  \vspace{3mm}

\textbf{Instruction:} You are given a science question (easy level) and four answer options (associated with “A”, “B”, “C”, “D”). Your task is to find the correct answer based on scientific facts, knowledge and reasoning. Don't generate anything other than one of the following characters: 'A B C D'.  \\   \vspace{1mm}

\textbf{Input:} Heavy forces on periodontal ligament causes: (A) Hyalinization (B) Osteoclastic activity around tooth (C) Osteoblastic activity around tooth (D) Crest bone resorption  \\   \vspace{1mm}

\textbf{Constraints:} The answer must be one or more of the following letters: 'A','B','C','D'. You must generate one and only one letter for each question. All questions have an answer. No justification is required.    \\   \vspace{1mm}

\textbf{Output:} 

\caption{Example of a 0-shot instruction used for the Multiple-Choice Question Answering (MCQA) task of MedMCQA dataset.}
\label{table:example-zero-shot-mcqa}
\end{table}

\newpage

\subsection{Method 2 - Few-shot}

In some cases, we mapped the original classes to more effective one's for each of the tasks, based on tries and errors (e.g: "entailment" has been map to "entails" for ChatGPT and Flan-T5 UL2 based on noticeable performances gains).

\begin{table}[h]
\small
\hrule  \vspace{3mm}
Prompt \\ \hrule  \vspace{3mm}

\textbf{Instruction:} You are given a science question (easy level) and four answer options (associated with “A”, “B”, “C”, “D”). Your task is to find the correct answer based on scientific facts, knowledge and reasoning. Don't generate anything other than one of the following characters: 'A B C D'.  \\   \vspace{1mm}

\textbf{Constraints:} The answer must be one or more of the following letters: 'A','B','C','D'. You must generate one and only one letter for each question. All questions have an answer. No justification is required.    \\   \vspace{1mm}

\textbf{Examples:}  \\   \vspace{1mm}

\textbf{Example 1:} Hyalinisation of the periodontal Ligament, due to excessive orthodontic forces results in (A) Frontal resorption (B) Undermining resorption (C) Cementum remaining intact (D) Dentine remaining intact 

\textbf{Output:} B  \\   \vspace{1mm}

\textbf{Example 2:} The earliest response of pulpitis is: (A) Cyst formation (B) Calcification (C) Hyalinization (D) Formation of dental granuloma 

\textbf{Output:} C  \\   \vspace{1mm}

\textbf{Example 3:} Among the secondary changes in tooth the most useful one for age determination is: (A) Attrition (B) Secondary dentine deposition (C) Root resorption (D) Root transparency 

\textbf{Output:} D  \\   \vspace{1mm}

\textbf{Example 4:} Feature of aging periodontium is (A) Lacunae in bone and cementum (B) Increased cell size (C) Increased cell number (D) Scalloping of cementum \& alveolar bone surface 

\textbf{Output:} D  \\   \vspace{1mm}

\textbf{Example 5:} Bacteria found in gingivitis are localized in (A) Connective tissue fibres (B) Gingival sulcus (C) Alveolar bone (D) Periodontal ligament 

\textbf{Output:} B  \\   \vspace{1mm}

\textbf{Input:} Heavy forces on periodontal ligament causes: (A) Hyalinization (B) Osteoclastic activity around tooth (C) Osteoblastic activity around tooth (D) Crest bone resorption 

\textbf{Output:} 

\caption{Example of a 5-shot instruction used for the Multiple-Choice Question Answering (MCQA) task of MedMCQA dataset.}
\label{table:example-few-shot-mcqa}
\end{table}

\newpage

\section{Relation Extraction}

\subsection{Method 1 - One-shot}

\begin{table}[h]
\small
\hrule  \vspace{3mm}
Prompt \\ \hrule  \vspace{3mm}

\textbf{Instruction:} Your goal is to do relation extraction and identifying if a gene-disease relation exist (positive) or not (negative). \\  

\textbf{Input :} These results suggest that the C1772T polymorphism in @GENE\$ is not involved in progression or metastasis of @DISEASE\$ \\   

\textbf{Constraints:} You have to output one label among « negative » or « positive ». Justification and explanations are prohibited. \\  

\textbf{Output:} \\

\caption{Example of a 0-shot instruction used for the Relation Extraction (RE) task of GAD dataset.}
\label{table:example-zero-shot-gad}
\end{table}


\subsection{Method 2 - Few-shot}

\begin{table}[h]
\small
\hrule  \vspace{3mm}
Prompt \\ \hrule  \vspace{3mm}

\textbf{Instruction:} Your goal is to do relation extraction and identifying if a gene-disease relation exist (positive) or not (negative).  \\  

\textbf{Constraints:} You have to output one label among « negative » or « positive ». Justification and explanations are prohibited.  \\ 

\textbf{Examples:}  \\  

\textbf{Example 1:} These findings suggest that the Gly460Trp polymorphism of @GENE\$ is not associated with @DISEASE\$. 

\textbf{Output:} Positive  \\  

\textbf{Example 2:} Our results suggest that deletion polymorphism of the @GENE\$ gene is not associated with the pathogenesis of @DISEASE\$ in Taiwanese. 

\textbf{Output:} Positive  \\ 

\textbf{Example 3:} The results suggest that the 5A/6A polymorphism of @GENE\$ gene may not be linked with appearance and/or progression of @DISEASE\$. 

\textbf{Output:} Positive  \\  

\textbf{Example 4:} Our study implies that the G/C polymorphism of the @GENE\$ gene may not be directly involved in the development and=or progression of @DISEASE\$. 

\textbf{Output:} Positive  \\ 

\textbf{Example 5:} Our study implies that the G/C polymorphism of the @GENE\$ gene may not be directly involved in the development and=or @DISEASE\$ of breast cancer.

\textbf{Output:} Negative  \\  

\textbf{Input:} These results suggest that the C1772T polymorphism in @GENE\$ is not involved in progression or metastasis of @DISEASE\$. 

\textbf{Output:} 


\caption{Example of a 5-shot instruction used for the Relation Extraction (RE) task of GAD dataset.}
\label{table:example-5-shot-gad}
\end{table}

\newpage

\section{Natural Language Inference}

\subsection{Method 1 - One-shot}


\begin{table}[h]
\small
\hrule  \vspace{3mm}
Prompt \\ \hrule  \vspace{3mm}

\textbf{Instruction:} Your goal is to do solve a natural language inference task by identifying if the hypothesis is either « entails » or « neutral » to the premise. \\   

\textbf{Input premise:} The liver is divided into the right lobe and left lobes. \\  

\textbf{Input hypothesis:} The gallbladder is near the right lobe of the liver.  \\   

\textbf{Constraints:} You have to output one label among « entails » or « neutral ». Justification and explanations are prohibited. \\  

\textbf{Output:} \\

\caption{Example of a 0-shot instruction used for the Natural Language Inference (NLI) task of SciTail dataset.}
\label{table:example-zero-shot-scitail}
\end{table}

\newpage

\subsection{Method 2 - Few-shot}

\begin{table}[h]
\small
\hrule  \vspace{3mm}
Prompt \\ \hrule  \vspace{3mm}

\textbf{Instruction:} Your goal is to do solve a natural language inference task by identifying if the hypothesis is either « entails » or « neutral » to the premise. \\   

\textbf{Constraints:} You have to output one label among « entails » or « neutral ». Justification and explanations are prohibited. \\ 

\textbf{Examples:} \\ 

\textbf{Example 1:}

\textbf{Premise:} Located primarily on the right side of the abdominal cavity, just above the duodenum, the liver aids in the digestion of fats by secreting bile into the duodenum.

\textbf{Hypothesis:} Most digestion is completed in the duodenum.

\textbf{Output:} neutral \\  

\textbf{Example 2:}

\textbf{Premise:} The brain is divided into the right and left hemisphere and each hemisphere is divided into 4 lobes called the frontal, temporal, occipital and parietal lobes.

\textbf{Hypothesis:} Each hemisphere of the cerebrum divided into 4 lobes.

\textbf{Output:} entails \\ 

\textbf{Example 3:}

\textbf{Premise:} The small intestine, where most digestion takes place, is a convoluted tube in the abdomen that begins at the pylorus of the stomach and ends at the opening to the large intestine.

\textbf{Hypothesis:} Most of the digestion reactions occur in the small intestine.

\textbf{Output:} entails \\ 

\textbf{Example 4:}

\textbf{Premise:} The small intestine is the long, thin segment of bowel that begins at the stomach and ends at the large intestine or colon.

\textbf{Hypothesis:} The small intestine begins in the stomach.

\textbf{Output:} entails \\  

\textbf{Example 5:}

\textbf{Premise:} The small intestine begins at the stomach and ends at the colon (large intestine).

\textbf{Hypothesis:} The small intestine begins in the stomach.

\textbf{Output:} entails \\ 

\textbf{Premise:} The liver is divided into the right lobe and left lobes. 

\textbf{Hypothesis:} The gallbladder is near the right lobe of the liver.  

\textbf{Output:} \\

\caption{Example of a 5-shot instruction used for the Natural Language Inference (NLI) task of SciTail dataset.}
\label{table:example-few-shot-scitail}
\end{table}

\newpage

\section{Classification}

\subsection{Method 1 - One-shot}

\begin{table}[h]
\small
\hrule  \vspace{3mm}
Prompt \\ \hrule  \vspace{3mm}

\textbf{Instruction:} Your goal is to do solve a classification task by identifying if one or more of the following hallmarks of cancer are present in the document: « evading growth suppressors », « tumor promoting inflammation », « enabling replicative immortality », « cellular energetics », « resisting cell death », « activating invasion and metastasis », « genomic instability and mutation », « none », « inducing angiogenesis », « sustaining proliferative signaling » or « avoiding immune destruction ». \\  

\textbf{Input:} Cytotoxicity was shown in manganese-treated groups ( 100 , 200 , 400 , and 800microM of MnCl(2) ) , and cell viability was decreased to 58.8\% of the control group at 2days after treatment with 800microM of MnCl(2) . \\   

\textbf{Constraints:} You have to output one or more label(s) among « evading growth suppressors », « tumor promoting inflammation », « enabling replicative immortality », « cellular energetics », « resisting cell death », « activating invasion and metastasis », « genomic instability and mutation », « none », « inducing angiogenesis », « sustaining proliferative signaling » or « avoiding immune destruction ». Justification and explanations are prohibited.

\textbf{Output:} 

\caption{Example of a 0-shot instruction used for the classification (CLS) task of HoC dataset.}
\label{table:example-zero-shot-hoc}
\end{table}

\newpage

\subsection{Method 2 - Few-shot}

\begin{table}[h]
\small
\hrule
\vspace{3mm}
Prompt \\ \hrule  \vspace{3mm}

\textbf{Instruction:} Your goal is to do solve a classification task by identifying if one or more of the following hallmarks of cancer are present in the document: « evading growth suppressors », « tumor promoting inflammation », « enabling replicative immortality », « cellular energetics », « resisting cell death », « activating invasion and metastasis », « genomic instability and mutation », « none », « inducing angiogenesis », « sustaining proliferative signaling » or « avoiding immune destruction ». \\

\textbf{Constraints:} You have to output one or more label(s) among « evading growth suppressors », « tumor promoting inflammation », « enabling replicative immortality », « cellular energetics », « resisting cell death », « activating invasion and metastasis », « genomic instability and mutation », « none », « inducing angiogenesis », « sustaining proliferative signaling » or « avoiding immune destruction ». Justification and explanations are prohibited. \\ 

\textbf{Examples:} \\ 

\textbf{Example 1:} However , significant cytotoxicity was only observed in PCB 52 concentrations larger than 0.1 microg ml(-1) , while there was no significant inhibition in PCB 77-treated cells at concentrations selected .

\textbf{Output:} none \\  

\textbf{Example 2:} In MeT-5A cells , both CNTs caused a dose-dependent induction of DNA damage ( \% DNA in comet tail ) in the 48-h treatment and SWCNTs additionally in the 24-h treatment , with a statistically significant increase at 40 \\u03bcg/cm(2) of SWCNTs and ( after 48 h ) 80 \\u03bcg/cm(2) of both CNTs .

\textbf{Output:} none \\  

\textbf{Example 3:} Copper-induced DNA strand breakage was first observed after 24 h of exposure , and was recorded again at 96 h , at a copper concentration of 20 microg l(-1) .

\textbf{Output:} genomic instability and mutation \\   

\textbf{Example 4:} Drug concentrations of 12.5 to 300 \u03bcM caused a pronounced reduction in cell survival rates five days after treatment , whereas concentrations higher than 25 \u03bcM were effective in reducing the survival rates to However , the maximum apoptosis frequency was 20.4\% for 25 \u03bcM cisplatin in cells analyzed at 72 h , indicating that apoptosis is not the only kind of cell death induced by cisplatin .

\textbf{Output:} none \\ 

\textbf{Example 5:} In contrast , in MCF 7 cells , molecular iodine ( 100 microM ) inhibited growth from 100\% to 83\% but delta-iodolactone ( 1 , 5 and 10 microM ) dose-dependently decreased growth rate from 100\% to 82\% and 62\% , respectively .

\textbf{Output:} none \\   

\textbf{Input:} Cytotoxicity was shown in manganese-treated groups ( 100 , 200 , 400 , and 800microM of MnCl(2) ) , and cell viability was decreased to 58.8\% of the control group at 2days after treatment with 800microM of MnCl(2) .

\textbf{Output:} 

\caption{Example of a few-shot instruction used for the classification (CLS) task of HoC dataset.}
\label{table:example-few-shot-hoc}
\end{table}

\newpage

\section{Semantic Textual Similarity}

\subsection{Method 1 - One-shot}

\begin{table}[h]
\small
\hrule
\vspace{3mm}
Prompt \\ \hrule

\vspace{3mm}

\textbf{Instruction:} Give me a similarity score beetween 0 et 5 and only the similarity score. \\   \vspace{1mm}


\textbf{Input:} The original sentence is : "- Eviter le contact de l'embout avec l'œil ou les paupières." can you tell me if the sentence is similar to : "Evitez le contact de l'embout du flacon avec l'œil ou les paupières.".  \\   \vspace{1mm}

\textbf{Output:} 

\caption{Example of a 0-shot instruction used for the Semantic Textual Similarity (STS) task of DEFT-2020 task 1 dataset.}
\label{table:example-zero-shot-deft-2020-t1}
\end{table}

\end{document}